\documentclass[conference]{IEEEtran}
\IEEEoverridecommandlockouts

\usepackage{cite}
\usepackage{amsmath,amssymb,amsfonts}
\usepackage{algorithmic}
\usepackage{graphicx}
\usepackage{textcomp}
\usepackage{xcolor}
\def\BibTeX{{\rm B\kern-.05em{\sc i\kern-.025em b}\kern-.08em
    T\kern-.1667em\lower.7ex\hbox{E}\kern-.125emX}}
    
\usepackage[acronyms,toc,nonumberlist,nogroupskip]{glossaries}
\glsdisablehyper
\glsnoexpandfields
\newacronym{ai}{AI}{artificial intelligence}
\newacronym{ad}{AD}{automated driving}
\newacronym[plural=DNNs,firstplural=deep neural networks (DNNs)]{dnn}{DNN}{deep neural network}
\newacronym{ilsvrc}{ILSVRC}{ImageNet large scale visual recognition challenge}
\newacronym{map}{mAP}{mean average precision}
\newacronym[plural=VRUs,firstplural=vunerable road users (VRU)]{vru}{VRU}{vunerable road user}

\begin{document}

\title{Towards Establishing Systematic Classification Requirements for Automated Driving\\
\thanks{{This work received funding from \emph{\mbox{VIVID}}, promoted by the German Federal Ministry of Education and Research, based on a decision of the Deutsche Bundestag with the grant number 16ME0173.}}
}

\author{\IEEEauthorblockN{1\textsuperscript{st} Ken T. Mori}
\IEEEauthorblockA{\textit{Institute of Automotive Engineering} \\
\textit{Technical University of Darmstadt}\\
Darmstadt, Germany \\
ken.mori@tu-darmstadt.de}
\and
\IEEEauthorblockN{2\textsuperscript{nd} Trent Brown}
\IEEEauthorblockA{\textit{Department of Mechanical Engineering} \\
\textit{Virginia Tech}\\
Blacksburg, United States of America \\
trentbrown@vt.edu}
\and
\IEEEauthorblockN{3\textsuperscript{rd} Steven Peters}
\IEEEauthorblockA{\textit{Institute of Automotive Engineering} \\
\textit{Technical University of Darmstadt}\\
Darmstadt, Germany \\
steven.peters@tu-darmstadt.de}
}

\IEEEpubid{\makebox[\columnwidth]{accepted for publication at IEEE IV 2023~\copyright2023 IEEE \hfill} \hspace{\columnsep}\makebox[\columnwidth]{ }}

\maketitle

\begin{abstract}
   Despite the presence of the classification task in many different benchmark datasets for perception in the automotive domain, few efforts have been undertaken to define consistent classification requirements.
   This work addresses the topic by proposing a structured method to generate a classification structure.
   First, legal categories are identified based on behavioral requirements for the vehicle.
   This structure is further substantiated by considering the two aspects of collision safety for objects as well as perceptual categories.
   A classification hierarchy is obtained by applying the method to an exemplary legal text.
   A comparison of the results with benchmark dataset categories shows limited agreement.
   This indicates the necessity for explicit consideration of legal requirements regarding perception.
\end{abstract}


\section{Introduction}
\label{sec:intro}

In the last decade, \glspl{dnn} have emerged in the field of \gls{ad}, where they have shown large progress on various tasks including perception~\cite{Huang.2020}.
Regarding perception, different tasks such as image classification, object detection and semantic segmentation are distinguished.
In each case, classification is performed for either the whole image, an object or an image pixel~\cite{Wu.10.08.2019}.
The classification task defines a set of classes for which the class likelihoods are estimated~\cite{Everingham.2010,Zhou.12.03.2021}.

At the same time, the requirements for classification remain unclear.
Since the rise of \gls{dnn}, research has been driven by the emergence of various datasets~\cite{Lin.01.05.2014}. 
These datasets specify the requirements of perception systems and are used in the verification process~\cite{Ashmore.10.05.2019}.
However, current datasets exhibit conflicting labeling taxonomies with inconsistent class definitions~\cite{Lambert.2020, Orsic.02.09.2020}. 
Moreover, there is no clear connection between the dataset categories and safety and legal requirements.
Such requirements are demanded by expected regulatory milestones such as the European \gls{ai} Act~\cite{EuropeanCommissionDirectorateGeneralforCommunicationsNetworksContentandTechnology.2021} or the proposed implementation for automated driving functions~\cite{EuropeanCommission.2022} of Regulation (EU) 2019/2144~\cite{CounciloftheEuropeanUnion.27.11.2019}.
This work therefore aims to specify classification requirements conforming to safety and legal requirements.

\section{Related Work}
\label{sec:related_work}

This section covers relevant datasets, ontologies and evaluation metrics from literature in the context of classification. 

\subsection{Dataset Categories}

Due to the large influence of datasets on perception~\cite{Lin.01.05.2014}, this section jointly considers different classification-relevant tasks and datasets with focus on the automotive domain.

For image classification, popular large-scale datasets may number thousands of categories~\cite{J.Deng.2009, Kuznetsova.2020}, while the number of categories is orders of magnitude smaller for the other tasks. 
2D object detection datasets such as~\cite{Everingham.2010, Lin.01.05.2014, Kuznetsova.2020} only include as many as 600 categories in the case of Open Images~\cite{Kuznetsova.2020}.

More specifically for the driving context, the maximum number of classes for object detection is 23 classes defined by nuScenes~\cite{Caesar.26.03.2019b}. 
This applies across 2D ~\cite{Geyer.14.04.2020, Yu.2020} and 3D ~\cite{Geiger.2012, Caesar.26.03.2019b, Sun.10.12.2019} detection datasets.
The number of categories considered for semantic segmentation ranges between 24-66~\cite{Cordts.06.04.2016, Huang.2020c, Yu.2020, Neuhold.2017}.
Other aspects such as lane markings~\cite{Huang.2020c, Yu.2020} or semantic maps ~\cite{Caesar.26.03.2019b, Houston.25.06.2020} used for segmentation of the drivable area~\cite{Pan.2020, Hendy.2020} are typically considered separately. 
The existence of stuff categories without object properties~\cite{Caesar.12.12.2016} leads to the higher number of categories compared to object detection.
Even within the same automotive domain, ambiguities and inconsistencies exist~\cite{Zendel.2018, Huang.2020c}, leading to limited class agreement across datasets.

\subsection{Ontologies for Classification}

An ontology provides a semantically expressive formal specification of concepts including entities and their relations~\cite{Klueck.2018}.
The general concept has been implicitly applied for datasets ~\cite{Cordts.06.04.2016, Lin.01.05.2014, Kuznetsova.2020} as well as for unifying taxonomies across different datasets~\cite{Meletis.15.03.2018}.
However, these approaches do not explicitly refer to ontologies and do not provide further reasoning for the semantic hierarchy.

An explicit use for general purpose classification is provided by ImageNet, which uses an ontology as basis for a semantic structure which is then populated with images~\cite{J.Deng.2009}.
It relies on WordNet, which provides a lexical database of concepts including synonyms as well as semantic relations between concepts~\cite{Miller.1995}.
ImageNet is also leveraged by YOLO9000 to scale up object detection regarding classification~\cite{Redmon.25.12.2016}.

Within traffic context, ontologies have been created for the purpose of generating test scenarios for \gls{ad} functions~\cite{Bagschik.29.03.2017, Klueck.2018}.
However, these approaches focus on driving behavior and do not explicitly consider perception or classification.
Other works unify classification taxonomies across domains for perception, but apply flat taxonomies for compatibility with standard training procedures~\cite{Lambert.2020, Zendel.2022, Bevandic.2022}.

\subsection{Perception Evaluation}

Common detection metrics such as \gls{map} define positive and negative samples per class ~\cite{Everingham.2010}, thereby considering the correct category as prerequisite for matching~\cite{Wu.10.08.2019}.
While some modifications to \gls{map} have been made for 3D detection in driving context~\cite{Caesar.26.03.2019b, Sun.10.12.2019}, this general procedure for matching has remained the same. 
Similar ideas are reflected in the classification accuracy as evaluated by the \gls{ilsvrc}, which adopts the Top-5 accuracy to account for label ambiguity~\cite{J.Deng.2009}.
These evaluation procedures all neglect the severity of different misclassifications.

Classification is also evaluated by loss functions which are used for optimization of classifiers.
The standard loss for classification is the cross-entropy loss, which also neglects the structure between classes~\cite{Wu.02.09.2017}.
However, proposals have been made in order to incorporate class similarity~\cite{Kobs.2020}, the severity of class confusion regarding semantic structure~\cite{Russakovsky.02.09.2014, Redmon.25.12.2016, Wu.02.09.2017} or regarding collision safety in traffic context~\cite{Liu.2020b}. 
More specifically, semantic similarity is conceptualized in a hierarchical classification structure either as distance metric~\cite{Russakovsky.02.09.2014, Redmon.25.12.2016, Wu.02.09.2017} or as depth of the lowest shared node~\cite{Russakovsky.02.09.2014}.
It should be noted that standard training methods rely on flat taxonomies~\cite{Lambert.2020}, while a hierarchical loss is difficult to optimize~\cite{Wu.02.09.2017}.
In addition, these approaches are either not designed for traffic context or do not consider the laws governing traffic.

\section{Method} 

This work presents a systematic method for developing a classification structure.
First, the objectives regarding the classification structure are discussed. 
This is followed by the overall method and each of its three steps.

\subsection{Objectives}

This section discusses the objectives regarding the classification structure and its attributes. 
In the classification structure, each node represents an object category.
This work considers objects to be nouns where at least one definition has ``object'' among its inherited hypernyms according to WordNet~\cite{Miller.1995}.
When designing an ontology, the constructed graph should be acyclic~\cite{Bagschik.29.03.2017} as in the tree structure of ImageNet~\cite{J.Deng.2009}.
This work therefore follows the approach of constructing such a hierarchical tree structure. 

Intuitively, class confusions may differ with respect to their severity~\cite{Liu.2020b}. 
For instance, confusing a tree with a pole may be inconsequential while confusing a person with a pole may be critical.
Therefore, it is desirable to systematically encode this severity in the structure.
Previous work focused on conceptualizing the severity of the class confusion as a symmetric distance in the classification hierarchy~\cite{Russakovsky.02.09.2014, Redmon.25.12.2016, Wu.02.09.2017}.
However, the direction of a class confusion may also impact road safety.
For example, confusing a child with a dog may be more severe than the reversed case when considering potential consequences. 
Capturing such relationships requires additional consideration beyond distance metrics.

\subsection{Overview}

Overall, the structure is built in three consecutive steps considering legal aspects, collision safety and perception.
Each step adds categories to the tree structure.

This work considers conformity to regulations as the first step in creating the classification structure.
Regulations demand different behavior towards different objects.
For example, a vehicle may be required to yield to traffic participants while this is not required for wild animals. 
Therefore, it is necessary to distinguish these objects accordingly.

The initial classification structure based on legal requirements does not explicitly distinguish different types of collisions. 
This work therefore introduces additional categories if they differ regarding collision safety. 
Both the aspect of collision severity~\cite{Volk.2020} and the aspect of collision likelihood~\cite{G.Nilsson.2004, Lefevre.2014} are considered in accordance with ISO 26262-1:2018~\cite{InternationalOrganizationforStandardization.122018}.

When considering legal and safety aspects, objects are categorized based on their behavioral requirements rather than by their appearance. 
This may lead to perceptually distinct objects such as trees and walls occupying the same category. 
Therefore, additional interpretable categories are introduced to facilitate perception and labeling.
For this purpose, categories relevant to humans~\cite{Lin.01.05.2014} as well as frequency and coverage ~\cite{Cordts.06.04.2016, Yu.2020} are considered.

\subsection{Legal Structure}

This section outlines the process of extracting categories from legal requirements. 

\subsubsection{Legal Basis}

While future regulations for perception of automated vehicles are unknown, current proposals for future regulations demand conformity with existing laws~\cite{EuropeanCommissionDirectorateGeneralforCommunicationsNetworksContentandTechnology.2021, EuropeanCommission.2022}. 
More generally, this work assumes that future regulations will be in a similar human readable format as existing regulations.
Therefore, the German StVO~\cite{BundesministeriumfurJustizundVerbraucherschutz.06.03.2013} along with a translation~\cite{Bottcher.2018} is arbitrarily selected as example to illustrate the method of this work. 
It is assumed that the ego vehicle is a car without special rights and below the weight of two tonnes above which further restrictions apply~\cite{BundesministeriumfurJustizundVerbraucherschutz.06.03.2013}.

\subsubsection{Extracting Requirements}

The general concept of using existing traffic laws as a basis for an ontology is similar to the ideas in~\cite{Bagschik.29.03.2017}. 
Another related approach is presented by the behavior-semantic scenery description which represents behavior-relevant requirements from the static scenery~\cite{Glatzki.2021}.
However, this work focuses on creating a unified hierarchical structure of all classification relevant objects instead of disjoint aspects limited to the static scenery. 

Specifically, this work first extracts all behavioral requirements for the ego vehicle as well as the respective preconditions. 
The internal behavior which includes the internal states of the system~\cite{Nolte.2017} is also considered. 
Some requirements are implicit, such as avoiding collisions with potentially harmful objects as well as perceiving objects. 
Since this work focuses on object categories, any relations between objects or attributes which may change over time are neglected.
Any objects mentioned in this context simply receive the requirement of perceiving them.
Examples of spatial relations are vehicle queues, diverging lanes or number of lanes on a carriageway, while attributes may include the state of a traffic light signal or whether a light is flashing.

\begin{figure}
  \centering
\includegraphics[width=\linewidth]{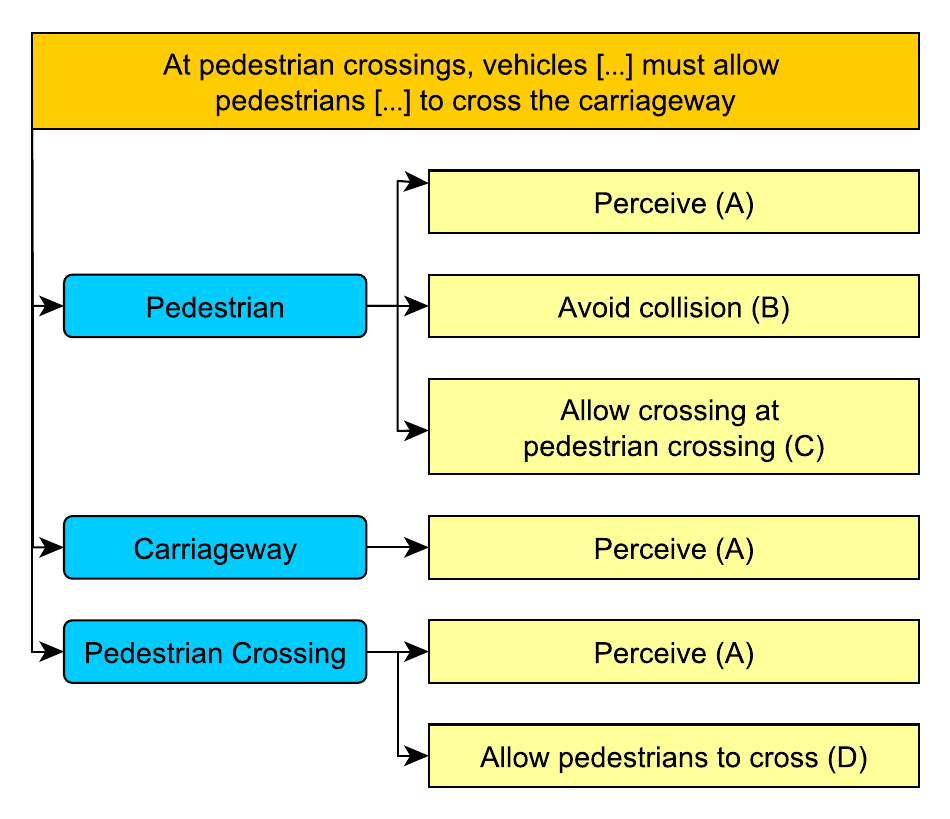}

   \caption{Example for process of grouping behavioral requirements by object categories. Uppercase letters are introduced as shorthand for visualization in Fig.~\ref{fig:structure_nodes}.}
   \label{fig:group_requirement}
\end{figure}

An example of this process shown in Fig.~\ref{fig:group_requirement}.
The original requirement that ``At pedestrian crossings, vehicles [...] must allow pedestrians [...] to cross the carriageway''~\cite[Sec.26 (1)]{Bottcher.2018} is decomposed into requirements regarding the different object categories. 
The carriageway is an example which itself does not require any specific action in this case, but requires perception to adhere to the behavioral requirement. 
The result is a list of object categories, each of which is associated with a set of behavioral requirements.

\begin{figure}
  \centering
\includegraphics[width=\linewidth]{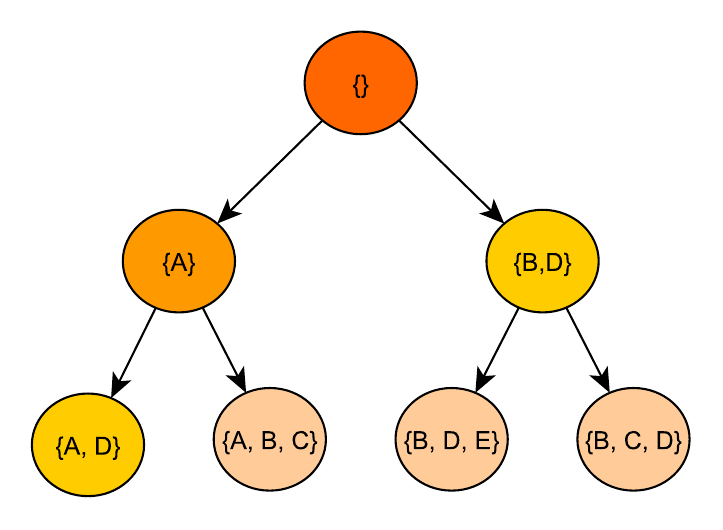}

   \caption{Example for iteratively adding nodes to structure. Stronger saturation and darker value indicate nodes added earlier.}
   \label{fig:structure_nodes}
\end{figure}

\subsubsection{Extracting Categories}

The categories grouping the requirements cannot be applied directly.
One issue is the existence of overlapping categories such as the vehicle category which includes buses.
Using the hyponyms in WordNet~\cite{Miller.1995}, the requirements of the general categories are added to the subcategories to resolve this issue. 
Additionally, it is possible that different categories have the same behavioral requirements. 
In this case, distinguishing them is not required from a legal perspective.
Accordingly, they are merged into categories which correspond to a unique set of behavioral requirements. 
For notational convenience, if a set of requirements is unique to one or few object categories, the name of the respective object categories are used as the node name. 

\subsubsection{Structuring Categories}

These extracted sets of requirements are subsequently structured in a hierarchical manner as visualized in Fig.~\ref{fig:structure_nodes}.
Each node shows an exemplary set of requirements, where each requirement is abbreviated by an uppercase letter.
The root node is considered to be objects for which no behavioral requirements exist, which corresponds to an empty set.
From there, nodes are iteratively added starting with the lowest number of requirements.
For each new node or set of requirements, all subsets within the tree are identified as parent node candidates. 
Among the parent node candidates within the same branch, the lowest level in the hierarchy is selected. 
More specifically, those candidates which are subsets of other parent node candidates are omitted.
If multiple candidates remain, the selection is based on the severity of infractions. 
Specifically, the relative complement of the new node and the parent node candidates is considered, which yields one set of behavioral requirements per parent node candidate.
For each of the sets, the severest penalty as defined in the penalty catalogue regulation (BKatV)~\cite{BundesministeriumfurJustizundVerbraucherschutz.14.03.2013} for disregarding the requirements in the set is considered. 
The set of requirements incurring the most severe penalty is selected as parent node.

\subsection{Safety Structure}

This section adds categories to the structure by considering the aspects of collision severity and collision likelihood.
Road regulations do not explicitly distinguish or rank collisions, but merely demand that no person should be harmed, endangered or inconvenienced~\cite{BundesministeriumfurJustizundVerbraucherschutz.06.03.2013}.

However, the severity can be distinguished by considering the penalties for infractions. 
When considering the penalties for breaking traffic laws as stated in~\cite{BundesministeriumfurJustizundVerbraucherschutz.14.03.2013} in accordance with~\cite{BundesministeriumfurJustizundVerbraucherschutz.06.03.2013, BundesministeriumfurJustizundVerbraucherschutz.05.03.2003}, the penalty typically increases from law infraction over hindering others to causing property damage.
Even harsher penalties may occur if bodily harm is caused by negligence, which is considered a criminal offence by German law~\cite{BundesministeriumfurJustizundVerbraucherschutz.13.11.1998}.
Road safety reports from the United Kingdom~\cite{DepartmentofTransport.2021} and Germany ~\cite{StatistischesBundesamt.25.07.2022} also distinguish between property damage and different degrees of injuries.

The accident rates with injuries are lower for cars compared to other road users~\cite{DepartmentofTransport.2021} due to the larger crush zones~\cite{Richards.2010}.
Other road users such as motorcyclists, pedestrians and bicyclists~\cite{DepartmentofTransport.2021} are often summarized as \glspl{vru}~\cite{Volk.2020}.
Therefore, this work follows the distinction between \gls{vru} and other vehicles.
Another aspect of severity for automated vehicles is the potential harm to passengers when colliding with obstacles.
For collisions with static objects in run-off-roadway accidents, the severity differs between different object categories~\cite{Lee.1999}.
Similarly, different animal species also differ in their respective accident severity~\cite{Huijser.082008}.
While predicting the severity of potential accidents is difficult, this work estimates whether property damage or injury is likely based on the rigidity and size of the object involved. 

While velocity impacts likelihood~\cite{G.Nilsson.2004} and severity of accidents~\cite{Han.2012, Richards.2010}, it is distinct from the object class even though the possible dynamic is influenced by the class~\cite{Dietmayer.2016}.
Since variations and speed limit violations may occur, this work does not consider velocity.
Another factor related to accident likelihood during driving is the occurrence of unexpected behaviors such as deviation from nominal behavior, conflict between traffic participants or violation of traffic laws~\cite{Lefevre.2014}. 
Some objects such as pedestrians are inherently more difficult to predict due to fast changes in movement speed or direction~\cite{Ahmed.2019}.
Other notable examples are animals and children, where erratic behavior is to be expected. 

These criteria are used to further subdivide the previously obtained categories where applicable.
Accident severity and accident likelihood are concurrent aspects which are not hierarchically ranked. 
However, avoiding collisions may not always be possible, especially for malicious movement of other traffic participants~\cite{ShalevShwartz.21.08.2017}. 
Therefore, this work places severity higher in the hierarchy than likelihood or erratic behavior.

\subsection{Perception Structure}

This section adds categories by considering perception with respect to interpretability, frequency and coverage. 

An obvious addition to the previously defined categories are object categories which are explicitly mentioned in the legal text~\cite{BundesministeriumfurJustizundVerbraucherschutz.06.03.2013} used as example in this work.
Additionally, popular benchmark dataset categories are included since they relate to perceptual aspects such as frequency~\cite{Cordts.06.04.2016}, diversity, coverage~\cite{Yu.2020} and visual similarity~\cite{Ertler.10.09.2019}. 
This work considers datasets for both object detection~\cite{Sun.10.12.2019, Geiger.2012, Geyer.14.04.2020, Yu.2020, Caesar.26.03.2019b} and semantic segmentation~\cite{Huang.2020c, Geyer.14.04.2020, Yu.2020, Neuhold.2017} in driving context.
All of these object categories are used as leaf node candidates. 

First, all leaf node candidates from the legal text are integrated into the class structure. 
This is followed by inserting the leaf node candidates originating from the datasets.
The placement according to the behavioral requirements is aided by considering synonyms and hypernyms of the leaf node candidates as defined by WordNet~\cite{Miller.1995}.
While the behavioral requirements can be deduced from the legal text, the categorization according to the safety structure requires human expert knowledge.
Semantically overlapping categories such as cars and taxis may coexist if their legal and safety requirements are identical.

\section{Results}

\begin{figure}[!b]
  \centering
    \includegraphics[width=\linewidth]{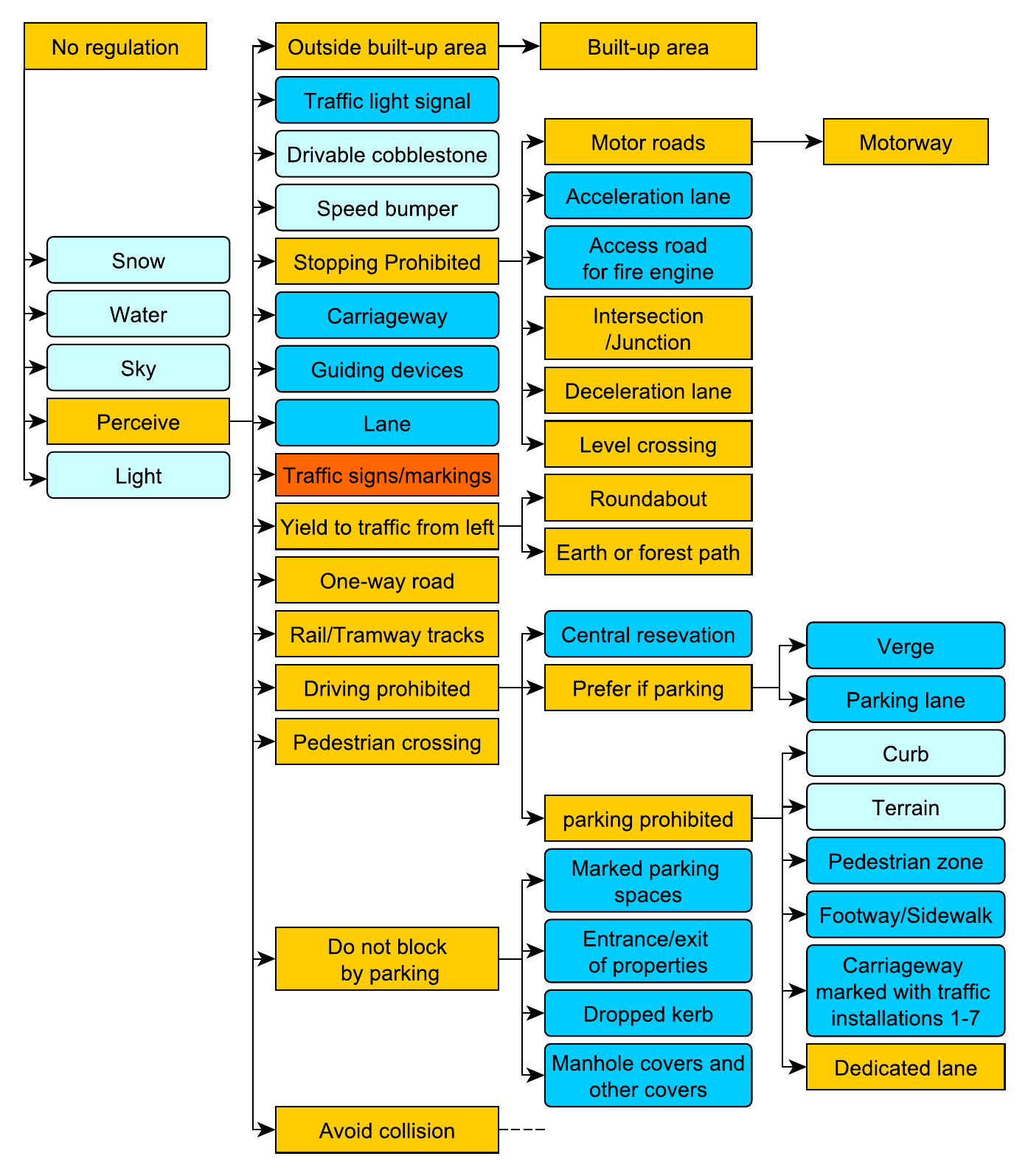}
   \caption{Classification hierarchy for non collision-relevant categories. 
   Yellow indicates legal categories, blue indicates perception categories from the legal text in dark blue and from datasets in light blue. Orange nodes contain further categories which are neglected in this work.}
   \label{fig:full_class_structure_1}
\end{figure}

\begin{figure*}
  \centering
    \includegraphics[width=\textwidth]{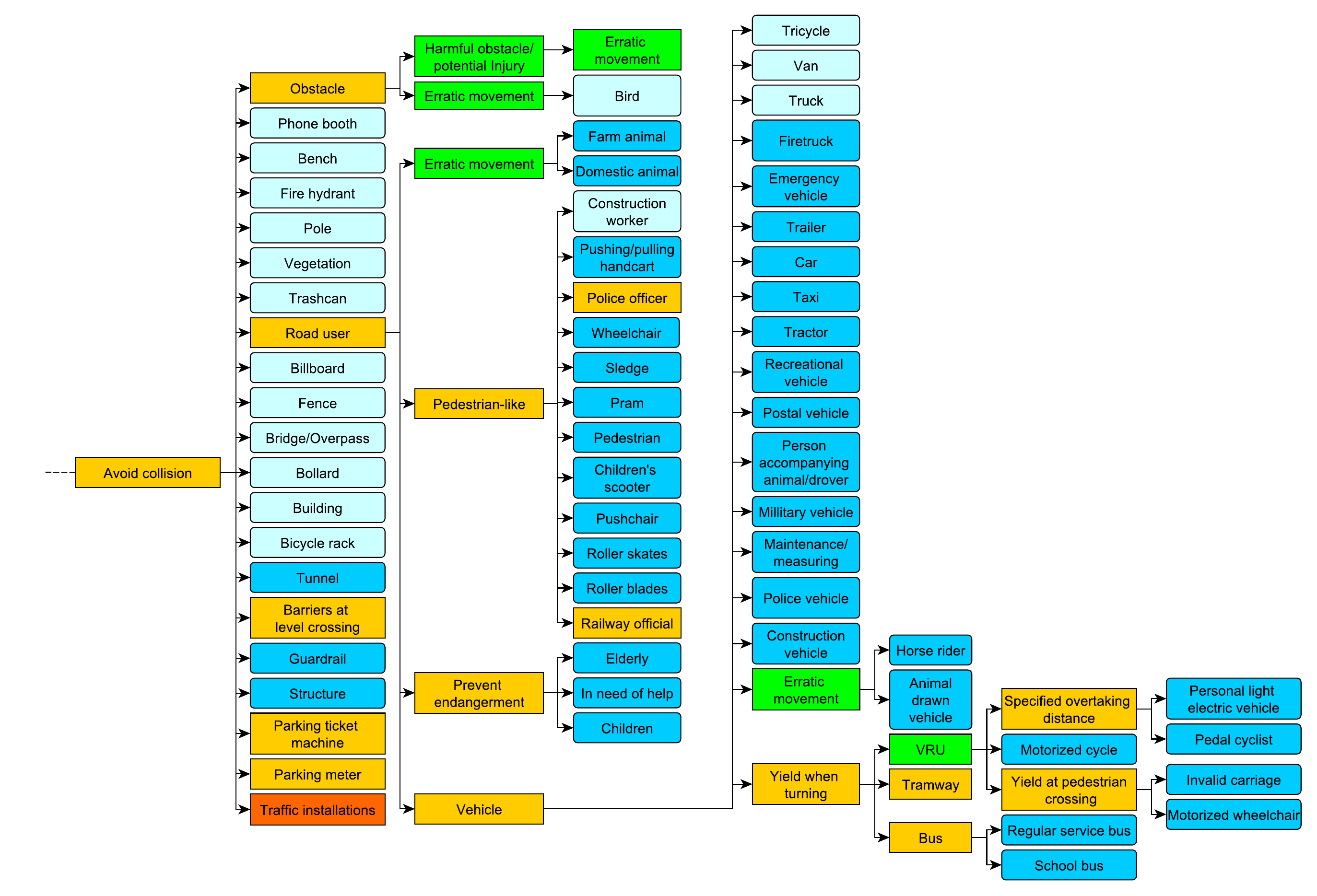}
   \caption{Classification hierarchy for collision-relevant objects. Continuation of the classification hierarchy for collision-relevant categories from Fig.~\ref{fig:full_class_structure_1} with identical  color coding. Green additionally indicates categories based on collision safety. Note that the vehicle categories were moved to the right for visualization purposes.
   Therefore, their visualized horizontal position does not directly correspond to their depth in the hierarchy.}
   \label{fig:full_class_structure_2}
\end{figure*}

This section presents the results of applying the presented method to create a classification structure.
The resulting hierarchy is split into Fig.~\ref{fig:full_class_structure_1} and Fig.~\ref{fig:full_class_structure_2} at the ``avoid collision'' node to improve visibility. 

\subsection{Legal Structure}

The results of the legal structure are presented in Fig.~\ref{fig:full_class_structure_1} and Fig.~\ref{fig:full_class_structure_2} by the yellow nodes.
The legal structure distinguishes 39 categories across seven levels when including the ``no regulation'' category.
Each node corresponds to a unique set of applicable legal requirements.
Object categories such as ``bus'' may be used for brevity of notation, but nevertheless represent the corresponding set of requirements.
Approximately half of the nodes correspond to non collision-relevant objects such as locations, while the collision relevant objects include both static objects and dynamic road users.
The two categories ``traffic signs/markings'' and ``traffic installations'' displayed in orange contain further categories which are not treated in this work.

\subsection{Safety Structure}

The categories related to safety add six nodes.
Therefore, they only have moderate impact on the overall structure as depicted by the green nodes in Fig.~\ref{fig:full_class_structure_2}.
The reason is that behavioral requirements from the legal structure already encode many differences between \gls{vru} and other traffic participants.
One modification is that obstacles are distinguished based on whether injury is to be expected from a potential collision.
Potential erratic movement such as exhibited by animals may be expected for road users, vehicles, and for both harmful obstacles and other obstacles.
Furthermore, the additional distinction of \gls{vru} increases the overall depth of the hierarchy to eight levels. 

\subsection{Perception Structure}

The categories obtained by considering interpretability and perception are visualized in Fig.~\ref{fig:full_class_structure_1} and Fig.~\ref{fig:full_class_structure_2} as blue nodes.
Dark blue categories originate from the legal text while light blue categories originate from datasets. 

Adding the categories from the legal text yields 16 nodes that are not collision-relevant and 39
collision-relevant nodes, summing up to 55 additional categories. 
The detection datasets add an another five categories while the segmentation datasets contribute 20 categories.
Overall, eight categories of the datasets are not collision-relevant and 17 categories are collision-relevant.
The majority of collision-relevant objects which the datasets contribute are static.

\section{Discussion} \label{sec:discussion}

This section presents a general discussion followed by a discussion on the severity of misclassifications.
After outlining some limitations of this work, results are compared with dataset categories and the potential impact on perception is discussed.

\subsection{General Discussion}

This work considers complementary categories obtained from considering legal, safety and perception requirements. 
The respective stakeholders are regulatory institutions, traffic participants and developers of automated vehicles.

Legal categories constitute the minimum required categories for compliance with regulations. 
Their benefit is their general applicability to any object encountered.
This includes rare categories typically considered as out-of-distribution without further distinction between different types of objects~\cite{Pinggera.2016, Geyer.14.04.2020, Blum.2019, Chan.2021}.
In addition, safety is improved by explicitly encoding it in the categories.
Further future substantiation may quantify likelihood and severity of collisions based on accident statistics.
Finally, the categories from perceptual considerations facilitate human interpretability and perception.
The results of this work are based on an argumentation and thus require further validation regarding their utility for the driving task.

\subsection{Severity of Misclassification}

While this work develops a hierarchical classification structure, distances between nodes only capture severity to a limited degree. 
One reason is that categories may relate to different aspects such as legal, safety and perception requirements which are difficult to compare. 
For legal categories, larger distances correspond to more legal transgressions. 
However, the severity of these transgressions depends on the specifics of each node. 

Legal categories with only one child node of a legal category are candidates for a potential conservative estimate. 
Overall, only four such cases exist.
Additionally, the incorrect classification of motorways or built-up areas leads to a severe misjudgement of the permissible velocity. 
Therefore, the potential hindrance or harm to other traffic participants is considered unacceptable, leaving only two conservative estimates. 
Overall, most classification errors between legal categories directly lead to a transgression of rules and are thus considered critical. 
This may indicate that hierarchical losses are not necessarily required.

\subsection{Limitations} \label{sec:discussion_legal} 

Currently, the legal structure neglects different types of traffic signs, markings and traffic installations.
The main reasons are limitations regarding attributes and relations.
Category-dependant attributes are relevant for the task of \gls{ad}~\cite{Metwaly.2021}.
A notable example is the dynamic state of traffic lights~\cite{Fregin.2018}.
In addition, resolving the interactions between traffic signs, traffic lights and police officers requires the state of traffic lights as well as relations between these objects~\cite{BundesministeriumfurJustizundVerbraucherschutz.06.03.2013}.
Some datasets including attributes already exist~\cite{Metwaly.2021}.
Other work has considered spatial and semantic relations between objects and regarding road and infrastructure~\cite{Buechel.2017, Karimi.2020} or structuring behavioral requirements of the static environment~\cite{Glatzki.2021}.
While incorporating attributes and relations into the ontology of this work is likely possible, this is left for future work.

Another limitation of this work is its application to a single legal text.
The general methodology presented in this work is applicable to any legal text.
Nevertheless, it is unclear to what degree the results agree when using a different text as basis.
Comparing results and potentially identifying a generalizable structure which applies across countries is left for future work.

\subsection{Comparison with Datasets}

Current dataset categories differ substantially from the categories defined in this work.
This indicates that current datasets do not sufficiently consider the regulatory and safety perspective of the driving task. 

The method in this work identifies 49 legal and safety categories and 100 perceptual categories.
This number is higher than the maximum number of 66 segmentation classes from Mapillary Vistas~\cite{Neuhold.2017} and 23 detection classes of nuScenes~\cite{Caesar.26.03.2019b}.
Furthermore, this work currently neglects different types of traffic signs, markings and traffic installations. 
Here, additional categories are expected, which may include as many as 313 traffic signs or more depending on which countries are considered~\cite{Ertler.10.09.2019}.
Some datasets already exist which include lane markings~\cite{Huang.2020c} or are specifically designed for traffic signs~\cite{Ertler.10.09.2019}.
Integrating these classes into the structure of this work is likely possible, but left for future work.

\subsection{Impact on Perception}

Some legal categories represent locations which are not boxable as considered by the object detection task.
This indicates the necessity for segmentation in environment perception which has so far received less attention in 3D space when compared to object detection.
The required semantic segmentation task may be considered as 3D semantic segmentation~\cite{Huang.2020c} or as segmentation of a birds-eye view semantic map~\cite{Caesar.26.03.2019b}.

For a higher number of categories, certain types of detection architectures such as class-specific convolutional heads are unsuitable regarding computation and speed~\cite{Zhou.12.03.2021, Singh.05.12.2017}.
Various architectures specifically for large-scale classification such as separate classification heads~\cite{Ertler.10.09.2019, Zhu.26.08.2019}, hierarchical extensions to softmax classification~\cite{Redmon.25.12.2016, Guo.30.05.2019} and expert models~\cite{Liu.17.03.2020, Niitani.25.10.2019} are proposed in literature. 
The classification requirements developed in this work may therefore require reconsideration regarding perception architectures.
Stronger classification requirements may especially impact the utility of different sensors in a fusion scheme since camera is better suited for classification than lidar or radar~\cite{Feng.2020}.

\section{Conclusion and Outlook}

This work considered defining perception requirements for classification.
To this end, a methodology for creating a structure based on legal requirements was developed and applied.
This structure was further substantiated with respect to collision safety.
Taking the legal text as well as benchmark datasets into account, the final perceptual categories were defined. 
The resulting hierarchy defines the required granularity in classification as well as conservative estimates.
The severity of class confusion is encoded to some extent. 

Future work could expand the structure with further requirements regarding attributes of objects as well as relations between different objects.
This is particularly relevant for the interaction of traffic signs, markings and light signals which were neglected in this work. 

This work shows the insufficiency of current perception datasets with respect to legal and safety requirements.
The new requirements impact the division between detection and segmentation classes, suitable perception algorithms and suitable sensor modalities. 
Further work is required to understand the effects of stricter classification requirements as well as finding suitable solutions. 
We hope that this work encourages explicit consideration of legal and safety requirements for classification and that it can serve as future reference.


\bibliographystyle{unsrt}
\bibliography{Mori.bib}

\begin{thebibliography}{10}

\bibitem{Huang.2020}
Yu~Huang and Y.~Chen.
\newblock Autonomous driving with deep learning: A survey of state-of-art
  technologies.
\newblock {\em ArXiv}, abs/2006.06091, 2020.

\bibitem{Wu.10.08.2019}
Xiongwei Wu, Doyen Sahoo, and Steven C.~H. Hoi.
\newblock Recent advances in deep learning for object detection.
\newblock {\em Neurocomputing}, 396:39--64, 2020.

\bibitem{Everingham.2010}
M.~Everingham, L.~{van Gool}, C.~K.~I. Williams, J.~Winn, and A.~Zisserman.
\newblock The pascal visual object classes (voc) challenge.
\newblock {\em International Journal of Computer Vision}, 88(2):303--338, 2010.

\bibitem{Zhou.12.03.2021}
Xingyi Zhou, Vladlen Koltun, and Philipp Kr{\"a}henb{\"u}hl.
\newblock Probabilistic two-stage detection.
\newblock {\em ArXiv}, abs/2103.07461, 2021.

\bibitem{Lin.01.05.2014}
Tsung-Yi Lin, Michael Maire, Serge Belongie, Lubomir Bourdev, Ross Girshick,
  James Hays, Pietro Perona, Deva Ramanan, C.~Lawrence Zitnick, and Piotr
  Doll{\'a}r.
\newblock Microsoft coco: Common objects in context.
\newblock In David Fleet, Tomas Pajdla, Bernt Schiele, and Tinne Tuytelaars,
  editors, {\em Computer Vision -- ECCV 2014}, pages 740--755, Cham, 2014.
  {Springer International Publishing}.

\bibitem{Ashmore.10.05.2019}
Rob Ashmore, Radu Calinescu, and Colin Paterson.
\newblock Assuring the machine learning lifecycle: Desiderata, methods, and
  challenges.
\newblock {\em ACM Computing Surveys}, 54(5):1--39, 2022.

\bibitem{Lambert.2020}
J.~Lambert, Z.~Liu, O.~Sener, J.~Hays, and V.~Koltun.
\newblock Mseg: A composite dataset for multi-domain semantic segmentation.
\newblock In {\em 2020 IEEE/CVF Conference on Computer Vision and Pattern
  Recognition (CVPR)}, pages 2876--2885, 2020.

\bibitem{Orsic.02.09.2020}
Marin Or{\v{s}}i{\'c}, Petra Bevandi{\'c}, Ivan Grubi{\v{s}}i{\'c}, Josip
  {\v{S}}ari{\'c}, and Sini{\v{s}}a {\v{S}}egvi{\'c}.
\newblock Multi-domain semantic segmentation with pyramidal fusion.
\newblock {\em ArXiv}, abs/2009.01636, 2020.

\bibitem{EuropeanCommissionDirectorateGeneralforCommunicationsNetworksContentandTechnology.2021}
{European Commission, Directorate-General for Communications Networks, Content
  and Technology}.
\newblock Proposal for a regulation of the european parliament and of the
  council laying down harmonised rules on artificial intelligence (artificial
  intelligence act) and amending certain union legislative acts, 2021.

\bibitem{EuropeanCommission.2022}
{European Commission}.
\newblock Annexes to the commission implementing regulation laying down rules
  for the application of regulation (eu) 2019/2144 of the european parliament
  and of the council as regards uniform procedures and technical specifications
  for the type-approval of the automated driving system (ads) of fully
  automated vehicles: Draft act: Annex 2, 2022.

\bibitem{CounciloftheEuropeanUnion.27.11.2019}
{Council of the European Union} and {European Parliament}.
\newblock Regulation (eu) 2019/2144 of the european parliament and of the
  council of 27 november 2019 on type-approval requirements for motor vehicles
  and their trailers, and systems, components and separate technical units
  intended for such vehicles, as regards their general safety and the
  protection of vehicle occupants and vulnerable road users, amending
  regulation (eu) 2018/858 of the european parliament and of the council and
  repealing regulations (ec) no 78/2009, (ec) no 79/2009 and (ec) no 661/2009
  of the european parliament and of the council and commission regulations (ec)
  no 631/2009, (eu) no 406/2010, (eu) no 672/2010, (eu) no 1003/2010, (eu) no
  1005/2010, (eu) no 1008/2010, (eu) no 1009/2010, (eu) no 19/2011, (eu) no
  109/2011, (eu) no 458/2011, (eu) no 65/2012, (eu) no 130/2012, (eu) no
  347/2012, (eu) no 351/2012, (eu) no 1230/2012 and (eu) 2015/166 (text with
  eea relevance): Regulation (eu) 2019/2144, 27.11.2019.

\bibitem{J.Deng.2009}
{J. Deng}, {W. Dong}, {R. Socher}, {L. -J. Li}, {Kai Li}, and {Li Fei-Fei}.
\newblock Imagenet: A large-scale hierarchical image database.
\newblock In {\em 2009 IEEE Conference on Computer Vision and Pattern
  Recognition}, pages 248--255, 2009.

\bibitem{Kuznetsova.2020}
Alina Kuznetsova, Hassan Rom, Neil Alldrin, Jasper Uijlings, Ivan Krasin, Jordi
  Pont-Tuset, Shahab Kamali, Stefan Popov, Matteo Malloci, Alexander
  Kolesnikov, Tom Duerig, and Vittorio Ferrari.
\newblock The open images dataset v4.
\newblock {\em International Journal of Computer Vision}, 128(7):1956--1981,
  2020.

\bibitem{Caesar.26.03.2019b}
Holger Caesar, Varun Bankiti, Alex~H. Lang, Sourabh Vora, Venice~Erin Liong,
  Qiang Xu, Anush Krishnan, Yu~Pan, Giancarlo Baldan, and Oscar Beijbom.
\newblock nuscenes: A multimodal dataset for autonomous driving.
\newblock In {\em 2020 IEEE/CVF Conference on Computer Vision and Pattern
  Recognition (CVPR)}, 2020.

\bibitem{Geyer.14.04.2020}
Jakob Geyer, Yohannes Kassahun, Mentar Mahmudi, Xavier Ricou, Rupesh Durgesh,
  Andrew~S. Chung, Lorenz Hauswald, Viet~Hoang Pham, Maximilian M{\"u}hlegg,
  Sebastian Dorn, Tiffany Fernandez, Martin J{\"a}nicke, Sudesh Mirashi,
  Chiragkumar Savani, Martin Sturm, Oleksandr Vorobiov, Martin Oelker,
  Sebastian Garreis, and Peter Schuberth.
\newblock A2d2: Audi autonomous driving dataset.
\newblock {\em ArXiv}, abs/2004.06320, 2020.

\bibitem{Yu.2020}
Fisher Yu, Haofeng Chen, Xin Wang, Wenqi Xian, Yingying Chen, Fangchen Liu,
  Vashisht Madhavan, and Trevor Darrell.
\newblock Bdd100k: A diverse driving dataset for heterogeneous multitask
  learning.
\newblock In {\em 2020 IEEE/CVF Conference on Computer Vision and Pattern
  Recognition (CVPR)}, 2020.

\bibitem{Geiger.2012}
Andreas Geiger, Philip Lenz, and Raquel Urtasun.
\newblock Are we ready for autonomous driving? the kitti vision benchmark
  suite.
\newblock In {\em Conference on Computer Vision and Pattern Recognition
  (CVPR)}, pages 3354--3361, 2012.

\bibitem{Sun.10.12.2019}
Pei Sun, Henrik Kretzschmar, Xerxes Dotiwalla, Aurelien Chouard, Vijaysai
  Patnaik, Paul Tsui, James Guo, Yin Zhou, Yuning Chai, Benjamin Caine, Vijay
  Vasudevan, Wei Han, Jiquan Ngiam, Hang Zhao, Aleksei Timofeev, Scott
  Ettinger, Maxim Krivokon, Amy Gao, Aditya Joshi, Sheng Zhao, Shuyang Cheng,
  Yu~Zhang, Jonathon Shlens, Zhifeng Chen, and Dragomir Anguelov.
\newblock Scalability in perception for autonomous driving: Waymo open dataset.
\newblock {\em ArXiv}, abs/1912.04838, 2019.

\bibitem{Cordts.06.04.2016}
Marius Cordts, Mohamed Omran, Sebastian Ramos, Timo Rehfeld, Markus Enzweiler,
  Rodrigo Benenson, Uwe Franke, Stefan Roth, and Bernt Schiele.
\newblock The cityscapes dataset for semantic urban scene understanding.
\newblock In {\em Proc. of the IEEE Conference on Computer Vision and Pattern
  Recognition (CVPR)}.

\bibitem{Huang.2020c}
Xinyu Huang, Peng Wang, Xinjing Cheng, Dingfu Zhou, Qichuan Geng, and Ruigang
  Yang.
\newblock The apolloscape open dataset for autonomous driving and its
  application.
\newblock {\em IEEE Transactions on Pattern Analysis and Machine Intelligence},
  42(10):2702--2719, 2020.

\bibitem{Neuhold.2017}
Gerhard Neuhold, Tobias Ollmann, Samuel~Rota Bul{\`o}, and Peter Kontschieder.
\newblock The mapillary vistas dataset for semantic understanding of street
  scenes.
\newblock In {\em 2017 IEEE International Conference on Computer Vision
  (ICCV)}, pages 5000--5009, 2017.

\bibitem{Houston.25.06.2020}
John Houston, Guido Zuidhof, Luca Bergamini, Yawei Ye, Long Chen, Ashesh Jain,
  Sammy Omari, Vladimir Iglovikov, and Peter Ondruska.
\newblock One thousand and one hours: Self-driving motion prediction dataset.
\newblock {\em ArXiv}, abs/2006.14480, 2020.

\bibitem{Pan.2020}
Bowen Pan, Jiankai Sun, Ho~Yin~Tiga Leung, Alex Andonian, and Bolei Zhou.
\newblock Cross-view semantic segmentation for sensing surroundings.
\newblock {\em IEEE Robotics and Automation Letters}, 5(3):4867--4873, 2020.

\bibitem{Hendy.2020}
Noureldin Hendy, Cooper Sloan, F.~Tian, Pengfei Duan, Nick Charchut, Y.~Xie,
  C.~Wang, and J.~Philbin.
\newblock Fishing net: Future inference of semantic heatmaps in grids.
\newblock {\em ArXiv}, abs/2006.09917, 2020.

\bibitem{Caesar.12.12.2016}
Holger Caesar, Jasper Uijlings, and Vittorio Ferrari.
\newblock Coco-stuff: Thing and stuff classes in context.
\newblock In {\em 2018 IEEE/CVF Conference on Computer Vision and Pattern
  Recognition}.

\bibitem{Zendel.2018}
Oliver Zendel, Katrin Honauer, Markus Murschitz, Daniel Steininger, and
  Gustavo~Fernandez Dominguez.
\newblock Wilddash - creating hazard-aware benchmarks.
\newblock In {\em Proceedings of the European Conference on Computer Vision
  (ECCV)}, 2018.

\bibitem{Klueck.2018}
F.~Klueck, Y.~Li, M.~Nica, J.~Tao, and F.~Wotawa.
\newblock Using ontologies for test suites generation for automated and
  autonomous driving functions.
\newblock In {\em 2018 IEEE International Symposium on Software Reliability
  Engineering Workshops (ISSREW)}, pages 118--123, 2018.

\bibitem{Meletis.15.03.2018}
Panagiotis Meletis and Gijs Dubbelman.
\newblock Training of convolutional networks on multiple heterogeneous datasets
  for street scene semantic segmentation.
\newblock In {\em 2018 IEEE Intelligent Vehicles Symposium (IV)}, pages
  1045--1050, Piscataway, NJ, 2018. IEEE.

\bibitem{Miller.1995}
George~A. Miller.
\newblock Wordnet: a lexical database for english.
\newblock {\em Communications of the ACM}, 38(11):39--41, 1995.

\bibitem{Redmon.25.12.2016}
Joseph Redmon and Ali Farhadi.
\newblock Yolo9000: Better, faster, stronger.
\newblock In {\em 2017 IEEE Conference on Computer Vision and Pattern
  Recognition}, pages 6517--6525.

\bibitem{Bagschik.29.03.2017}
Gerrit Bagschik, Till Menzel, and Markus Maurer.
\newblock Ontology based scene creation for the development of automated
  vehicles.
\newblock In {\em 2018 IEEE Intelligent Vehicles Symposium (IV)}, 2018.

\bibitem{Zendel.2022}
Oliver Zendel, Matthias Sch{\"o}rghuber, Bernhard Rainer, Markus Murschitz, and
  Csaba Beleznai.
\newblock Unifying panoptic segmentation for autonomous driving.
\newblock In {\em 2022 IEEE/CVF Conference on Computer Vision and Pattern
  Recognition (CVPR)}, pages 21319--21328, 2022.

\bibitem{Bevandic.2022}
Petra Bevandic, Marin Orsic, Ivan Grubisic, Josip Saric, and Sinisa Segvic.
\newblock Multi-domain semantic segmentation with overlapping labels.
\newblock {\em 2022 IEEE/CVF Winter Conference on Applications of Computer
  Vision (WACV)}, pages 2422--2431, 2022.

\bibitem{Wu.02.09.2017}
Cinna Wu, Mark Tygert, and Yann LeCun.
\newblock A hierarchical loss and its problems when classifying
  non-hierarchically.
\newblock {\em PloS one}, 14(12), 2019.

\bibitem{Kobs.2020}
Konstantin Kobs, M.~Steininger, Albin Zehe, Florian Lautenschlager, and
  A.~Hotho.
\newblock Simloss: Class similarities in cross entropy.
\newblock In Denis Helic, Gerhard Leitner, Martin Stettinger, Alexander
  Felfernig, and Zbigniew~W. Ra{\'s}, editors, {\em Foundations of Intelligent
  Systems}, pages 431--439, Cham, 2020. {Springer International Publishing}.

\bibitem{Russakovsky.02.09.2014}
Olga Russakovsky, Jia Deng, Hao Su, Jonathan Krause, Sanjeev Satheesh, Sean Ma,
  Zhiheng Huang, Andrej Karpathy, Aditya Khosla, Michael Bernstein,
  Alexander~C. Berg, and Li~Fei-Fei.
\newblock Imagenet large scale visual recognition challenge.
\newblock {\em International Journal of Computer Vision}, 115(3):211--252,
  2015.

\bibitem{Liu.2020b}
Xiao-Feng Liu, Yimeng Zhang, Xiongchang Liu, Song Bai, Site Li, and J.~You.
\newblock Reinforced wasserstein training for severity-aware semantic
  segmentation in autonomous driving.
\newblock In {\em 2020 IEEE/CVF Conference on Computer Vision and Pattern
  Recognition (CVPR)}, volume abs/2008.04751, 2020.

\bibitem{Volk.2020}
Georg Volk, J{\"o}rg Gamerdinger, Alexander~von Betnuth, and O.~Bringmann.
\newblock A comprehensive safety metric to evaluate perception in autonomous
  systems.
\newblock {\em 2020 IEEE 23rd International Conference on Intelligent
  Transportation Systems (ITSC)}, pages 1--8, 2020.

\bibitem{G.Nilsson.2004}
{G. Nilsson}.
\newblock {\em Traffic Safety Dimensions and the Power Model to Describe the
  Effect of Speed on Safety}.
\newblock Doctoral thesis, {Lund Institute of Technology}, 2004.

\bibitem{Lefevre.2014}
St{\'e}phanie Lef{\`e}vre, Dizan Vasquez, and Christian Laugier.
\newblock A survey on motion prediction and risk assessment for intelligent
  vehicles.
\newblock {\em ROBOMECH Journal}, 1(1):1--14, 2014.

\bibitem{InternationalOrganizationforStandardization.122018}
{International Organization for Standardization}.
\newblock Iso 26262-1:2018: Road vehicles - functional safety - part 1:
  Vocabulary, 12/2018.

\bibitem{BundesministeriumfurJustizundVerbraucherschutz.06.03.2013}
{Bundesministerium f{\"u}r Justiz und Verbraucherschutz}.
\newblock Stra{\ss}enverkehrs-ordnung: Stvo, 06.03.2013.

\bibitem{Bottcher.2018}
Lorenz B{\"o}ttcher.
\newblock Road traffic regulations (stra{\ss}enverkehrs-ordnung, stvo) with
  annexes, 2018.
\newblock https://germanlawarchive.iuscomp.org/?p=1290.

\bibitem{Glatzki.2021}
Felix Glatzki, Moritz Lippert, and Hermann Winner.
\newblock Behavioral attributes for a behavior-semantic scenery description
  (bssd) for the development of automated driving functions.
\newblock {\em 2021 IEEE International Intelligent Transportation Systems
  Conference (ITSC)}, pages 667--672, 2021.

\bibitem{Nolte.2017}
Marcus Nolte, Gerrit Bagschik, Inga Jatzkowski, Torben Stolte, Andreas Reschka,
  and Markus Maurer.
\newblock Towards a skill- and ability-based development process for self-aware
  automated road vehicles.
\newblock In {\em IEEE ITSC 2017}, pages 1--6, Piscataway, NJ, 2017. IEEE.

\bibitem{BundesministeriumfurJustizundVerbraucherschutz.14.03.2013}
{Bundesministerium f{\"u}r Justiz und Verbraucherschutz}.
\newblock Verordnung {\"u}ber die erteilung einer verwarnung, regels{\"a}tze
  f{\"u}r geldbu{\ss}en und die anordnung eines fahrverbotes wegen
  ordnungswidrigkeiten im stra{\ss}enverkehr (bu{\ss}geldkatalog-verordnung -
  bkatv): Bkatv, 14.03.2013.

\bibitem{BundesministeriumfurJustizundVerbraucherschutz.05.03.2003}
{Bundesministerium f{\"u}r Justiz und Verbraucherschutz}.
\newblock Stra{\ss}enverkehrsgesetz: Stvg.
\newblock {\em Bundesgesetzblatt}, 2003 Teil I(Nr. 10):310--344, 05.03.2003.

\bibitem{BundesministeriumfurJustizundVerbraucherschutz.13.11.1998}
{Bundesministerium f{\"u}r Justiz und Verbraucherschutz}.
\newblock Strafgesetzbuch: Stgb.
\newblock {\em Bundesgesetzblatt}, 1998 Teil I(Nr. 75):3322--3410, 13.11.1998.

\bibitem{DepartmentofTransport.2021}
Reported road causalties great britain, annual report: 2020: National
  statistics, 2021.
\newblock
  https://www.gov.uk/government/statistics/reported-road-casualties-great-britain-annual-report-2020/reported-road-casualties-great-britain-annual-report-2020\#overall-casualties.

\bibitem{StatistischesBundesamt.25.07.2022}
{Statistisches Bundesamt}.
\newblock Verkehr: Verkehrsunf{\"a}lle: April 2022.
\newblock
  https://www.destatis.de/DE/Themen/Gesellschaft-Umwelt/Verkehrsunfaelle/Publikationen/Downloads-Verkehrsunfaelle/verkehrsunfaelle-monat-2080700221044.pdf?\_\_blob\=publicationFile.

\bibitem{Richards.2010}
D.~C. Richards.
\newblock Relationship between speed and risk of fatal injury: Pedestrians and
  car occupants, 2020.

\bibitem{Lee.1999}
Jinsun Lee and Fred Mannering.
\newblock Analysis of roadside accident frequency and severity and roadside
  safety management, 1999.
\newblock https://depts.washington.edu/trac/bulkdisk/pdf/475.1.pdf.

\bibitem{Huijser.082008}
M.~P. Huijser, P.~McGowen, J.~Fuller, A.~Hardy, A.~Kociolek, A.~P. Clevenger,
  D.~Smith, and R.~Ament.
\newblock Wildlife-vehicle collision reduction study: Report to congress, 2008.
\newblock https://www.fhwa.dot.gov/publications/research/safety/08034/02.cfm.

\bibitem{Han.2012}
Yong Han, Jikuang Yang, Koji Mizuno, and Yasuhiro Matsui.
\newblock Effects of vehicle impact velocity, vehicle front-end shapes on
  pedestrian injury risk.
\newblock {\em Traffic injury prevention}, 13(5):507--518, 2012.

\bibitem{Dietmayer.2016}
Klaus Dietmayer.
\newblock Predicting of machine perception for automated driving.
\newblock In Markus Maurer, J.~Christian Gerdes, Barbara Lenz, and Hermann
  Winner, editors, {\em Autonomous Driving: Technical, Legal and Social
  Aspects}, pages 407--424. {Springer Berlin Heidelberg}, Berlin, Heidelberg,
  2016.

\bibitem{Ahmed.2019}
Sarfraz Ahmed, M.~Nazmul Huda, Sujan Rajbhandari, Chitta Saha, Mark Elshaw, and
  Stratis Kanarachos.
\newblock Pedestrian and cyclist detection and intent estimation for autonomous
  vehicles: A survey.
\newblock {\em Applied Sciences}, 9(11):38, 2019.

\bibitem{ShalevShwartz.21.08.2017}
Shai Shalev-Shwartz, Shaked Shammah, and Amnon Shashua.
\newblock On a formal model of safe and scalable self-driving cars.
\newblock {\em ArXiv}, abs/1708.06374, 2017.

\bibitem{Ertler.10.09.2019}
Christian Ertler, Jerneja Mislej, Tobias Ollmann, Lorenzo Porzi, Gerhard
  Neuhold, and Yubin Kuang.
\newblock The mapillary traffic sign dataset for detection and classification
  on a global scale.
\newblock In {\em Computer Vision - ECCV 2020}.

\bibitem{Pinggera.2016}
Peter Pinggera, Sebastian Ramos, Stefan Gehrig, Uwe Franke, Carsten Rother, and
  Rudolf Mester.
\newblock Lost and found: detecting small road hazards for self-driving
  vehicles.
\newblock In {\em 2016 IEEE/RSJ International Conference on Intelligent Robots
  and Systems (IROS)}, pages 1099--1106, 2016.

\bibitem{Blum.2019}
Hermann Blum, Paul-Edouard Sarlin, Juan Nieto, Roland Siegwart, and Cesar
  Cadena.
\newblock Fishyscapes: A benchmark for safe semantic segmentation in autonomous
  driving.
\newblock In {\em Proceedings of the IEEE/CVF International Conference on
  Computer Vision (ICCV) Workshops}, 2019.

\bibitem{Chan.2021}
Robin Chan, Krzysztof Lis, Svenja Uhlemeyer, Hermann Blum, Sina Honari, Roland
  Siegwart, Mathieu Salzmann, Pascal Fua, and Matthias Rottmann.
\newblock Segmentmeifyoucan: A benchmark for anomaly segmentation.
\newblock {\em ArXiv}, abs/2104.14812, 2021.

\bibitem{Metwaly.2021}
Kareem~M. Metwaly, Aerin Kim, Elliot Branson, and Vishal Monga.
\newblock Car - cityscapes attributes recognition a multi-category attributes
  dataset for autonomous vehicles.
\newblock {\em ArXiv}, abs/2111.08243, 2021.

\bibitem{Fregin.2018}
Andreas Fregin, Julian Muller, Ulrich Krebel, and Klaus Dietmayer.
\newblock The driveu traffic light dataset: Introduction and comparison with
  existing datasets.
\newblock In {\em 2018 IEEE International Conference on Robotics and Automation
  (ICRA)}, pages 3376--3383, 2018.

\bibitem{Buechel.2017}
M.~Buechel, G.~Hinz, F.~Ruehl, H.~Schroth, C.~Gyoeri, and A.~Knoll.
\newblock Ontology-based traffic scene modeling, traffic regulations dependent
  situational awareness and decision-making for automated vehicles.
\newblock In {\em 2017 IEEE Intelligent Vehicles Symposium (IV)}, pages
  1471--1476, 2017.

\bibitem{Karimi.2020}
A.~Karimi and P.~S. Duggirala.
\newblock Formalizing traffic rules for uncontrolled intersections.
\newblock In {\em 2020 ACM/IEEE 11th International Conference on Cyber-Physical
  Systems (ICCPS)}, pages 41--50, 2020.

\bibitem{Singh.05.12.2017}
Bharat Singh, Hengduo Li, Abhishek Sharma, and Larry~S. Davis.
\newblock R-fcn-3000 at 30fps: Decoupling detection and classification.
\newblock In {\em 2018 IEEE/CVF Conference on Computer Vision and Pattern
  Recognition (CVPR 2018)}, pages 1081--1090, Piscataway, NJ, 2018. IEEE.

\bibitem{Zhu.26.08.2019}
Benjin Zhu, Zhengkai Jiang, Xiangxin Zhou, Zeming Li, and Gang Yu.
\newblock Class-balanced grouping and sampling for point cloud 3d object
  detection.
\newblock {\em ArXiv}, abs/1908.09492, 2019.

\bibitem{Guo.30.05.2019}
Ye~Guo, Yali Li, and Shengjin Wang.
\newblock Cs-r-fcn: Cross-supervised learning for large-scale object detection.
\newblock In {\em ICASSP 2020 - 2020 IEEE International Conference on
  Acoustics, Speech and Signal Processing (ICASSP)}.

\bibitem{Liu.17.03.2020}
Yu~Liu, Guanglu Song, Yuhang Zang, Yan Gao, Enze Xie, Junjie Yan, Chen~Change
  Loy, and Xiaogang Wang.
\newblock 1st place solutions for openimage2019 -- object detection and
  instance segmentation.
\newblock {\em ArXiv}, abs/2003.07557, 2020.

\bibitem{Niitani.25.10.2019}
Yusuke Niitani, Toru Ogawa, Shuji Suzuki, Takuya Akiba, Tommi Kerola, Kohei
  Ozaki, and Shotaro Sano.
\newblock Team pfdet's methods for open images challenge 2019.
\newblock {\em ArXiv}, abs/1910.11534, 2019.

\bibitem{Feng.2020}
D.~Feng, C.~Haase-Sch{\"u}tz, L.~Rosenbaum, H.~Hertlein, C.~Gl{\"a}ser,
  F.~Timm, W.~Wiesbeck, and K.~Dietmayer.
\newblock Deep multi-modal object detection and semantic segmentation for
  autonomous driving: Datasets, methods, and challenges.
\newblock {\em IEEE Transactions on Intelligent Transportation Systems}, pages
  1--20, 2020.

\end{thebibliography}

\end{document}